\documentclass[sigconf, screen]{acmart}

\AtBeginDocument{%
  }
\usepackage{times}
\usepackage{latexsym}
\usepackage{booktabs} 
\usepackage{multirow}
\usepackage[normalem]{ulem}
\usepackage{enumitem}
\usepackage{subcaption}
\usepackage{algorithm}
\usepackage{tabularx}
\usepackage{algpseudocode}
\usepackage{array}
\usepackage{amsmath}
\usepackage{amssymb}
\usepackage{pgfplots}
\pgfplotsset{compat=1.18}
\usepackage{hyperref}
\hypersetup{
    colorlinks=true,
    linkcolor=blue,
    filecolor=magenta,      
    urlcolor=cyan,
    pdftitle={Overleaf Example},
    pdfpagemode=FullScreen,
    }

\setcopyright{acmlicensed}
\copyrightyear{2025}
\acmYear{2025}
\acmDOI{XXXXXXX.XXXXXXX}
\acmConference[Conference acronym 'XX]{Make sure to enter the correct
  conference title from your rights confirmation email}{October 27--31,
  2025}{Dublin, Ireland}
\acmISBN{978-1-4503-XXXX-X/2018/06}

\acmSubmissionID{5286}

\begin{document}

\settopmatter{printacmref=false}
\renewcommand\footnotetextcopyrightpermission[1]{}

\title{Adaptation Method for Misinformation Identification}


\author{Yanping Chen}
\affiliation{%
  \institution{Soochow University}
  \country{China}}

\author{Weijie Shi}
\affiliation{%
  \institution{Hong Kong University of Science and Technology}
  \country{China}
}

\author{Mengze Li}
\affiliation{%
 \institution{Hong Kong University of Science and Technology}
 \country{China}
}

\author{Yue Cui}
\affiliation{%
  \institution{Hong Kong University of Science and Technology}
  \country{China}
}

\author{Hao Chen}
\affiliation{%
  \institution{Tencent}
  \country{China}
}

\author{Jia Zhu}
\affiliation{%
  \institution{Zhejiang Normal University}
  \country{China}
}

\author{Jiajie Xu}
\affiliation{%
  \institution{Soochow University}
  \country{China}
}


\begin{abstract}

Multimodal fake news detection plays a crucial role in combating online misinformation. Unfortunately, effective detection methods rely on annotated labels and encounter significant performance degradation when domain shifts exist between training (source) and test (target) data. To address the problems, we propose ADOSE, an Active Domain Adaptation (ADA) framework for multimodal fake news detection which actively annotates a small subset of target samples to improve detection performance. To identify various deceptive patterns in cross-domain settings, we design multiple expert classifiers to learn dependencies across different modalities. These classifiers specifically target the distinct deception patterns exhibited in fake news, where two unimodal classifiers capture knowledge errors within individual modalities while one cross-modal classifier identifies semantic inconsistencies between text and images. To reduce annotation costs from the target domain, we propose a least-disagree uncertainty selector with a diversity calculator for selecting the most informative samples. The selector leverages prediction disagreement before and after perturbations by multiple classifiers as an indicator of uncertain samples, whose deceptive patterns deviate most from source domains. It further incorporates diversity scores derived from multi-view features to ensure the chosen samples achieve maximal coverage of target domain features. The extensive experiments on multiple datasets show that ADOSE outperforms existing ADA methods by 2.72\% $\sim$ 14.02\%, indicating the superiority of our model. 
\end{abstract}




\maketitle

\section{INTRODUCTION}

\begin{figure}[htbp]
    \begin{minipage}{0.42\textwidth}
        \centering
        \includegraphics[width=\linewidth]{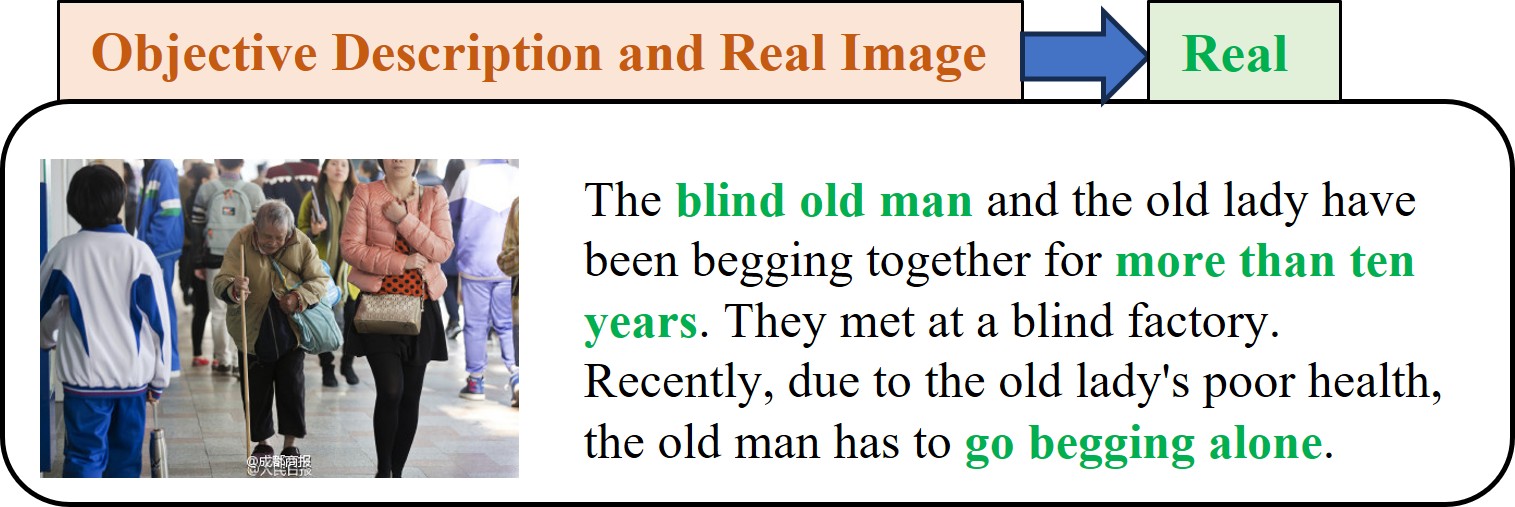}
        \vskip 0.2cm
        \includegraphics[width=\linewidth]{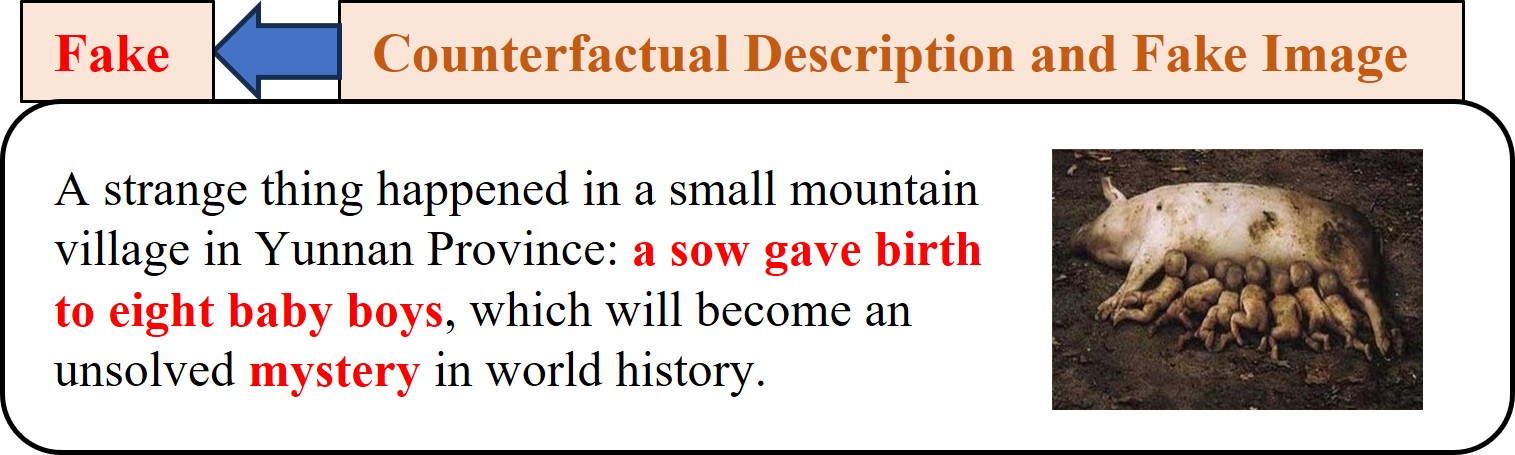}
        \subcaption{Society Domain}\label{fig:cha1and2}
    \end{minipage}
    

    \begin{minipage}{0.42\textwidth}
    \centering
    \begin{minipage}{0.48\linewidth} 
        \centering
        \includegraphics[width=\linewidth]{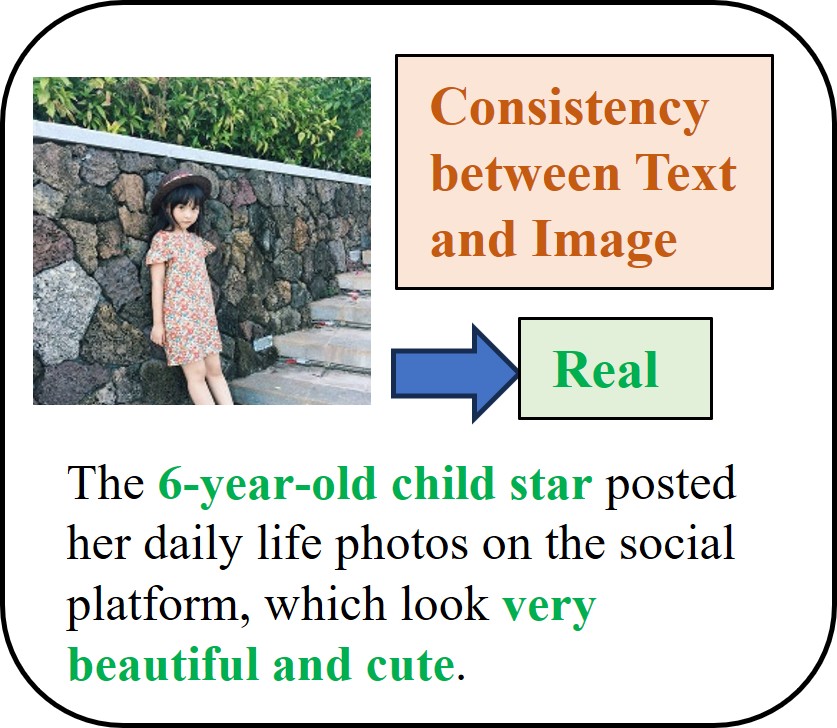}
    \end{minipage}%
    \hfill 
    \begin{minipage}{0.48\linewidth} 
        \centering
        \includegraphics[width=\linewidth]{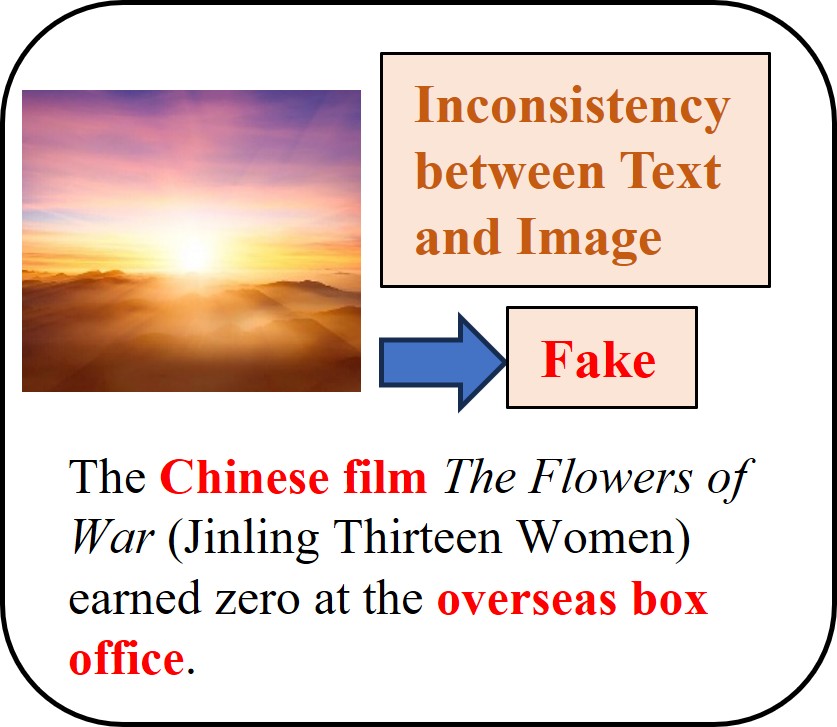}
    \end{minipage}%
    \subcaption{Entertainment Domain}\label{fig:cha3and4} 
    \end{minipage}%
    
    \caption{The comparison of deception patterns between the social and entertainment domains in the Weibo dataset, with "Real" labeled as real news and "Fake" labeled as false news.}
    \label{fig:challenge}
\end{figure}

Multimodal fake news detection aims to identify the authenticity of online news composed of multiple modalities such as text and images. With the proliferation of various social networks like Twitter and Weibo, this task has garnered significant attention due to its critical role in combating misleading information \cite{Mitigating, Unidistill, NLcentered}.

In the real world, unverified breaking news is particularly prone to misleading the public, necessitating effective fake news detection. Although supervised learning approaches have proven effective for fake news detection, they require extensive manual annotation of emerging news samples, which is time-consuming and labor-intensive \cite{Anomaly, lv2025grasp, li2023winner, li2023unsupervised, zhang2020devlbert}. Another category of domain adaptation methods transfers knowledge from labeled source domains to unlabeled target domains, but struggles with performance degradation when confronting significant domain shifts in data distribution \cite{UDArumor, onestage}. 
The expensive cost of label acquisition and the practical demands of domain adaptation have driven researchers to seek a balance between them, resulting in the rise of Active Domain Adaptation (ADA) methods \cite{AUDA}. ADA assists knowledge transfer in the target domain by strategically annotating informative target samples within a limited budget. Inspired by this idea, we introduce ADA for fake news detection in the domain adaptation setting when label acquisition of the target domain is difficult. 

Despite the success of ADA in various multimedia tasks such as object detection and semantic segmentation \cite{ADAobject, ADAsemantic}, it faces two key challenges in multimodal fake news detection across domains. \textbf{First, varying deception patterns across different domains complicate modality dependency learning}. Fake news in different domains often exhibits distinct deception patterns \cite{nlsm}, including knowledge errors within a modality (intra-modal dependency) and semantic inconsistencies between modalities (inter-modal dependency) \cite{event-radar, jointly}. In the social domain, as shown in Figure~\ref{fig:cha1and2}, deception patterns based on the intra-modal dependency manifest as counterfactual descriptions and fabricated images. In the entertainment domain, however, the inter-modal dependency is more prominent, characterized by inconsistencies between text and image content, as seen in Figure~\ref{fig:cha3and4}. In cross-domain settings, simultaneously capturing both intra- and inter-modal dependencies becomes crucial for learning knowledge from multiple domains.  
\textbf{Second, informative sample selection in the target domain should consider both domain shift and multimodal features}. An effective selection strategy requires identifying samples that reflect domain shifts most, which simultaneously characterize both the distinctive features and deceptive patterns of fake news in the target domain \cite{sharpness, Bilevel, LDM}. These shift samples typically facilitate faster distribution adaptation in the target domain. It also requires maximizing coverage of the overall target samples in multimodal feature space and minimizing redundancy among chosen samples. However, the complexity of domain shifts and multimodal features makes it challenging to design such an active selection strategy.


In this paper, we propose ADOSE, an Active Domain Adaptation Framework for Multimodal Fake News Detection. (1) For the complicated modality dependencies, We develop a Modal-dependency Expertise Fusion Network (MEFN) that separately learns intra- and inter-modal dependencies knowledge across domains and fuses all knowledge to identify deception patterns for fake news detection. We construct two unimodal classifiers to learn intra-modal dependencies and one cross-modal classifier to learn inter-modal dependencies. Adversarial networks and contrastive training are used to generate domain-invariant and semantically aligned features as input for the classifiers \cite{RDCM, coolant}. (2) To select informative samples on the target domain, we design a dual selection strategy targeting domain shift and feature distribution. Specifically, we utilize prediction uncertainty as an indicator of domain shifts and propose a Least-disagree Uncertainty Selector (LUS) to identify the samples exhibiting maximal predictive uncertainty. LUS applies Gaussian perturbations to measure a sample's proximity to decision boundaries as uncertainty estimation. Samples with higher uncertainty not only better exhibit the deceptive patterns of target news samples, but also contain more modality dependencies knowledge that benefits model learning. Then, we develop a Multi-view Diversity Calculator (MDC) to further select the most discriminative samples in the latent feature space based on diversity scores derived from multiple feature views. Through this process, we identify the most valuable target domain samples for annotation. Extensive experiments on multiple datasets demonstrate that ADOSE significantly outperforms existing ADA methods. The main contributions of this paper are as follows:

\begin{itemize} [topsep=7pt]
    \item We introduce ADOSE, an active domain adaptation framework for multimodal fake news detection, aiming to address the domain adaptation problem when labeled target domain samples are insufficient.
    \item We develop a Modal-dependency Expertise Fusion Network (MEFN) that simultaneously captures both intra-modal and inter-modal dependencies through multiple expert classifiers based on adversarial training and contrastive learning, enabling fake news pattern recognition across domains.
    \item We design a dual selection strategy that characterizes the contribution of target samples to domain adaptation through a least-disagree uncertainty metric obtained via Gaussian perturbation, while incorporating diversity scores derived from multi-view features to identify the most informative samples for annotation.
\end{itemize}

\section{RELATED WORK}
\textbf{Multimodal Fake News Detection.}
Deep learning \cite{li2024panoptic, li2023art, li2020unsupervised, zhang2022magic, miao2019pose, miao2021vspw} has demonstrated increasingly powerful data analysis capabilities. 
Among them, one of the fields that deserves more attention is multimodal technology \cite{ji2023binary, lin2025healthgpt, wu2024semantic, li2023fine, li2023variational, li2022fine}.
Multimodal fake news detection has received significant attention in recent years, with a core challenge of the effective fusion between text and image to improve detection performance. Current research has proposed various approaches to address multimodal fusion. For instance, COOLANT \cite{coolant} employs cross-modal contrastive learning to achieve more accurate image-text alignment. It incorporates an auxiliary task to soften the loss of negative samples and utilizes an attention mechanism to effectively integrate unimodal representations and cross-modality correlations. MSACA \cite{MSACA} focuses on multi-scale semantic alignment by constructing hierarchical multi-scale image representations and leveraging an attention module to select the most discriminative embeddings, enhancing semantic consistency between text and images. MMDFND \cite{MMDFND} targets multi-domain scenarios, integrating domain embeddings and stepwise transformer networks to address inter-domain modal semantic and dependency deviations, achieving domain-adaptive fusion. Although the above research has made some progress, multimodal information alignment \cite{zhang2022boostmis, zhang2019frame, zhang2021consensus,  miao2023temporal, lin2024non, lin2024action} still needs further research.


\textbf{Active Domain Adaptation.}
Active Domain Adaptation (ADA) seeks to enhance model generalization to new target domains by actively selecting and annotating a small, highly informative subset of target samples, which is an important method to prompt the development of representation learning \cite{dai2024mpcoder, dai2025less, li2022dilated}. Recent studies address the limitations of traditional Active Learning (AL) methods, which often overlook domain shift, by introducing innovative strategies tailored for ADA. For instance, CLUE \cite{clue} employs uncertainty-weighted clustering to select target samples that are both uncertain and diverse in feature space, while EADA \cite{eada} leverages an energy-based sampling approach, integrating domain characteristics and instance uncertainty to bridge the source-target gap. Similarly, DUC \cite{duc} introduces Dirichlet-based uncertainty calibration and a two-round selection strategy to mitigate miscalibration under distribution shift and Detective \cite{detective} extends ADA to multi-source settings using a dynamic DA model and evidential deep learning to handle complex domain gaps.


\section{PROBLEM STATEMENT}
Formally, we have access to \( M \) fully labeled source domains and a target domain with actively labeled target data within a pre-defined budget \( \mathcal{B} \). All domains are defined based on different news events or topics. The \( i \)-th source domain \( \mathcal{S}_{i} = \{ (t_{s,i}^{n}, v_{s,i}^{n}), y_{s,i}^{n} \}_{n=1}^{N_{s,i}} \) contains \( N_{s,i} \) labeled samples, where \( i \in M \) and the target domain \( \mathcal{T}_{u} = \{ (t_{tu}^{n}, v_{tu}^{n}) \}_{n=1}^{N_{tu}} \) contains \( N_{tu} \) unlabeled samples. And, \( (t, v) \) is a text-image pair, where \(t\) is a text sentence, and \(v\) is the corresponding image. Following the standard ADA setting \cite{eada}, the size of budget \( \mathcal{B} \) set to \( N_{tl} \), the labeled target domain is denoted as \( \mathcal{T}_{l} = \{ (t_{tl}^{n}, v_{tl}^{n}), y_{tl}^{n} \}_{n=1}^{N_{tl}} \), where \( N_{tl} \ll N_{tu} \) and \( N_{tl} \ll N_{s.i} \). The multiple source domains and target domain have the same label space \( \mathcal{Y} = \{ 0,1 \} \) (0 indicates real news and 1 indicates fake news). We aim to learn a model \( \mathcal{M}(\cdot) \) adapting from \( \{\mathcal{S}_{i}\}_{i=1}^{M} \) to \( \mathcal{T}_u \), i.e., the model can generalize well on unseen samples from \( \mathcal{T}_u \). In general, \( \mathcal{M}(\cdot) \) consists of two functions:
\begin{align}
&\underbrace{\mathcal{M}(\cdot; (\Theta_m, \Theta_a))}_{\text{ADA Model}} := \underbrace{\mathcal{M}_m((\mathcal{S}_{i})_{i=1}^{M}, \mathcal{T}_l); \Theta_m)}_{\text{Multi-domain Adaptation Model}} \notag \\
&\quad \rightarrow \underbrace{\mathcal{M}_a((\mathcal{T}_l |  \mathcal{T}_u); \Theta_a)}_{\text{Active Learning Model}}, \label{eq:1}
\end{align}
where \( \mathcal{M}_m(\cdot; \Theta_m) \) is the multi-domain learning model and \( \mathcal{M}_a(\cdot; \Theta_a) \) is the annotation candidate selection through active learning model with parameters \( \Theta_m \) and \( \Theta_a \).

\section{METHODOLOGY}
\begin{figure*}
    \centering
    \includegraphics[width=\textwidth]{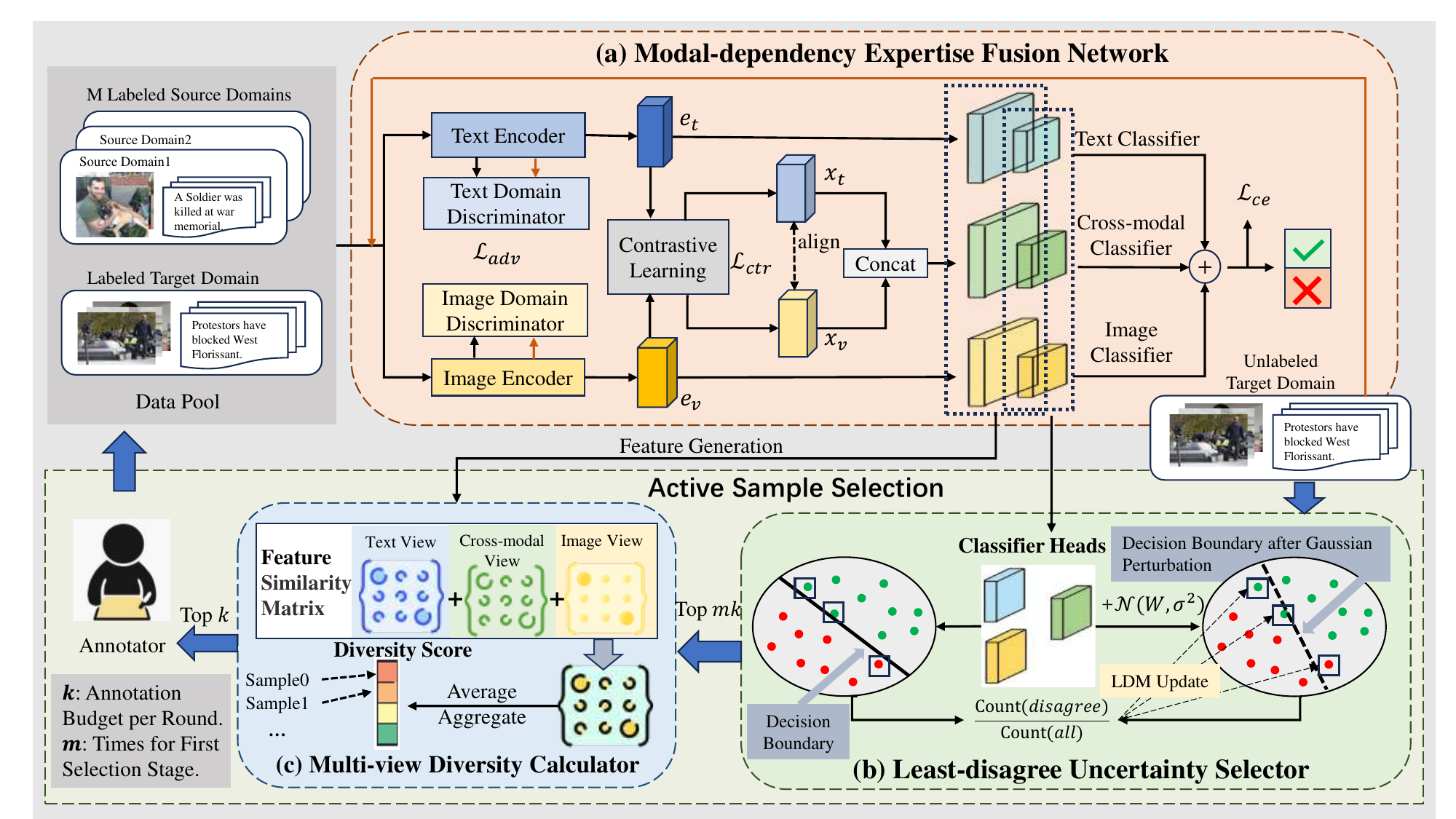}
    \caption{The network architecture of ADOSE. (a) Modal-dependency Expertise Fusion Network (MEFN) utilizes multiple classifiers to detect fake news based on adversarial training and contrastive learning. (b) and (c) are the Least-disagree Uncertainty Selector (LUS) and Multi-view Diversity Calculator (MDC) that respectively measure the uncertainty (first selection stage) and diversity (second selection stage) of target samples.}
    \label{fig:ADOSE}
\end{figure*}

In this section, we introduce our model, ADOSE, which comprises three components: Modal-dependency Expertise Fusion Network (MEFN), Least-disagree Uncertainty Selector (LUS) and Multi-view Diversity Calculator (MDC). The structure of the model is illustrated in Figure~\ref{fig:ADOSE}. Given a text-image pair from different domains, we first extract domain invariant features from both single-modal and cross-modal views and then integrate expertise from multiple classifiers for fake news detection by MEFN module (\S4.1). During active selection round, we quantify model's prediction uncertainty by LUS module (\S4.2) and compute the diversity score of target domain samples by MDC module (\S4.3) to select informative samples to annotate for domain adaptation.  


\subsection{Modal-dependency Expertise Fusion Network}
The Modal-dependency Expertise Fusion Network (MEFN) leverages multiple expertise learned from modality dependencies to construct a domain adaptation model for multimodal fake news classification, as shown in Figure~\ref{fig:ADOSE}(a). First, textual and image features from news in different domains are extracted into an invariant latent space using the corresponding encoders with discriminators (\S4.1.1).
Then, we use contrastive learning to align the semantic gap between text and images, thereby generating cross-modal feature representations (\S4.1.2). Based on the domain invariant features obtained from the three modalities, we separately construct classifiers for each modality and integrate the predictions of all classifiers for the final detection (\S4.1.3).

\subsubsection{Domain Invariant Feature Extraction.}
Given an input text-image pair \( (t,v) \) in each domain, following previous work \cite{pairfollow, RDCM}, we leverage a convolutional neural network (i.e., TextCNN \cite{CNN}) with an additional perceptron (MLP) as the textual encoder \( E_t \) to obtain the representation of \(t\) as \( e_t \). For image representation, following existing methods \cite{imagefollow1,imagefollow2}, we use ResNet50 as the visual backbone neural network and choose the feature of the final pooling layer as the initial visual embedding \(h_v\). Then, similar to the text modality, we also apply a MLP to further generate visual representation With reduced dimensionality of \( v \) as \( e_v \):
\begin{equation}
e_t = E_t(t; \theta_t), \quad e_v = E_v(v; \theta_v)
\end{equation}
where \( e_t, e_v \in \mathbb{R}^d \), \( E_v \) represents the visual encoder, and \( \theta_t, \theta_v \) represents the parameter of the textual encoder and the visual encoder respectively.

The feature distribution shift among source samples and target samples is a key obstacle for domain adaptation. In order to generate domain invariant features, we train the textual encoder and visual encoder through adversarial networks, which has been proven effective in recent domain adaptation studies \cite{dapm, SAUE}. Since our task setting involves \(M+1\) domains with different distributions, we design a textual discriminator \( D_t \) (for invariant text features) and a visual discriminator \( D_v \) (for invariant image features), each with \( K \) classes (\( K=M+1 \)). We perform fine-grained domain adversarial learning, where textual adversarial loss \( \mathcal{L}_{abv\_t} \) and visual adversarial loss \( \mathcal{L}_{abv\_v} \) are defined between any two different domains:
\begin{equation}
\mathcal{L}_{abv\_t} = - \mathbb{E}_{e_t\sim(S\ \cup\ \mathcal{T}_{u})} \mathrm{log} D_t(d_l\ |\ GRL(e_t))
\end{equation}
\begin{equation}
\mathcal{L}_{abv\_v} = - \mathbb{E}_{e_v\sim(S\ \cup\ \mathcal{T}_{u})} \mathrm{log} D_v(d_l\ |\ GRL(e_v))
\end{equation}
where \( d_l \) represent the domain labels of \(e_t\) and \(e_v\), with the domain label space defined as \(\{0,1,2...M\}\) (corresponding to \(M\) source domains and one target domain), and GRL denote the gradient reversal layer. Here, \( D_t(d_l\ |\ GRL(e_t)) \) and \( D_v(d_l\ |\ GRL(e_v)) \) specifically represent the probabilities predicted by the discriminators for the ground-truth domain label. We minimize the feature generator and maximize the overall adversarial loss which can be defined as \( \mathcal{L}_{adv} = \mathcal{L}_{abv\_t} + \mathcal{L}_{abv\_t} \) to play the min-max game.

\subsubsection{Multimodal Representation Alignment.}
Multimodal representation alignment is used to prepare for acquiring expert knowledge of modal interactions by aligning semantically consistent text and images. Based on the domain invariant text feature \( e_t \) and image feature \( e_v \), we utilize MLP transformations to further obtain textual and visual alignment representation \( x_{t} \) and \( x_{v} \):
\begin{equation}
    x_{t} = \text{MLP}(e_t; \theta_{mt}), \quad x_{v} = \text{MLP}(e_v; \theta_{mv})
\end{equation}

In the task of fake news detection, cross-modal correspondence or similarity is more likely to only exist in real news rather than in misinformation scenarios. Besides, texts for different misinformation examples may use the same image in a specific event domain, which results in the image and text of many negative samples being close to each other in the semantic space. To overcome this limitation, we adopt the sampling strategy proposed in RDCM \cite{RDCM}, which only takes real posts as positive samples and filters out negative samples with high semantic similarity on the visual modality with a weighting function as follows:
\begin{equation}
\begin{aligned}
\mathbb{I} \big( (e_t^i, e_v^i), (e_t^j, e_v^j) \big)) &= 
\begin{cases} 
0, & \text{if } \text{sim}(h_v^i, h_v^j) \geq \beta \\
\beta - \text{sim}(h_v^i, h_v^j), & \text{else}
\end{cases} \\[1ex]
\text{sim}(h_v^i, h_v^j) &= \left( \frac{h_v^i {h_v^j}^T}{\|h_v^i\| \|h_v^j\|} + 1 \right) / 2
\end{aligned}
\label{eq:similarity}
\end{equation}
Here \(i\) and \(j\) denote indices corresponding to the \(i\)-th and \(j\)-th samples in a batch, \(h_v^i\) and \(h_v^j\) denote the initial visual embedding of feature processing on \(e_v^i\) and \(e_v^j\) respectively (We specifically take \(h_v\) as the output of the softmax layer of the backbone for the visual modality (e.g., ResNet50 in our model)). And \( \text{sim}(h_v^i, h_v^j) \) represents the similarity between \( (e_t^i, e_v^i) \) and \( (e_t^j, e_v^j) \), and \( \beta \) is a threshold to remain semantic dissimilar pairs as negative samples. 

Then, we leverage the contrastive loss objective to update the parameters of MLP i.e. \( \theta_{mt} \) and \( \theta_{mv} \) so that we obtain multimodal semantic alignment features (i.e., \(x_t\) and \(x_v\)) which are utilized for inter-modal fusion in the following section. The contrastive objective is defined as:
\begin{equation}
    \mathcal{L}_{ctr} = -\mathrm{log} \frac{\mathrm{exp}\big(\frac{x_t^i {x_v^i}^T}{\tau}\big)}
    {\mathrm{exp}\big(\frac{x_t^i {x_v^i}^T}{\tau}\big) + \sum\limits_{i \neq j}^{b} \mathrm{exp}\big(\frac{x_t^i {x_v^j}^T}{\tau}\big) \mathbb{I} \big( (e_t^i, e_v^i), (e_t^j, e_v^j) \big))}
\end{equation}
where \(i\) represents the indices of real posts in a minibatch, \(j\) represents the indices of the other samples in this minibatch except the \(i\)-th sample, \(b\) is the minibatch size, and \(\tau\) is a temperature hyperparameter.

\subsubsection{Expertise Fusion Network.}
After the stages of domain invariant feature extraction and multimodal representation alignment, we obtain domain invariant unimodal feature representations \( e_t, e_v \) and aligned modal representations \( x_t, x_v \). We concatenate \( x_t \) and \( x_v \) to form the feature representation of the cross-model. Based on the unimodal feature representations, we construct a text classifier \( cls\_t(e_t;) \) and an image classifier \( cls\_v(e_v;) \) to model the intra-modal dependencies, respectively. Based on the cross-modal features, we construct a cross-modal classifier \( cls\_c([x_t, x_v];) \) to model the inter-modal dependencies. These all classifiers are implemented as two-layer MLP. Finally, we integrate the outputs of the three classifiers to form the predictions of the expert knowledge fusion network. This process is formulated as follows:
\begin{equation}
    score_y = \log P_{cls\_t}(y|e_t) + \log P_{cls\_v}(y|e_v) + \log P_{cls\_c}(y|[x_t, x_v]) \notag
\end{equation}
\begin{equation}
    P_{mefn} = \mathrm{softmax}\left(\frac{score_y}{\mathrm{log} \sum_{y \in \mathcal{Y}} \mathrm{exp}(score_y)}\right)
\end{equation}
where \(y\) denotes the news label (0 or 1), \( P_{cls\_t}, P_{cls\_v}, P_{cls\_c} \) represent the predicted probability distributions from three classifiers based input features, and \( P_{mefn} \) is the final probability distribution from our MEFN module.

\subsection{Least-disagree Uncertainty Selector}

The erroneous or uncertain decisions of the model on the target domain can reflect the extent of distribution shift between the target domain and the source domain samples. Selecting the most representative misclassified or low-confidence samples with limited annotation costs can help the model more effectively adapt to the unseen distribution of the target domain. Since we cannot directly determine when misclassification will occur, we consider the samples where the model is most uncertain, as these samples are more likely to be misclassified. Intuitively, samples closest to the decision boundary are considered the most uncertain. Inspired by recent active learning work \cite{LDM, LDM2014}, we develop a Least-disagree Uncertainty Selector (LUS) (Figure~\ref{fig:ADOSE}(b)) by introducing the Least Disagree Metric (LDM) to quantify the closeness of a sample to the decision boundary. Let \( \mathcal{X} = \{(e_t, e_v, x_t, x_v)\} \) and \( \mathcal{Y} \) denote the feature and label spaces, respectively, and \( \mathcal{H} \) denote the hypothesis space, encompassing all possible functions \( h: \mathcal{X} \to \mathcal{Y} \). The disagree metric between two hypotheses \( h_1 \) and \( h_2 \) is defined as follows:
\begin{equation}
    \rho(h_1, h_2) := \mathbb{P}_{\mathcal{X} \sim \mathcal{T}_{u}} [h_1(\mathcal{X}) \neq h_2(\mathcal{X})]
\end{equation}
where \( \mathbb{P}_{\mathcal{X} \sim \mathcal{T}_u} \) is the probability measure on \(\mathcal{X}\) induced by \( \mathcal{T}_u \). For a given hypothesis \( g \in \mathcal{H} \) and \( \mathcal{X}_0 \in \mathcal{X} \), let \( \mathcal{H}_{g, \mathcal{X}_0} := \{ h \in \mathcal{H} | h(\mathcal{X}_0) \neq g(\mathcal{X}_0) \} \) be the set of hypotheses disagreeing with \( g \) in their prediction for \( \mathcal{X}_0 \) and all the hypotheses belong to a parametric family. Further, the least disagree metric (LDM) is defined as
\begin{equation}
    L(g, \mathcal{X}_0) := \inf_{h \in \mathcal{H}_{g,\mathcal{X}_0}} \rho(h, g).
\end{equation}

However, the true LDM in our task setting is not computable because the complex multimodal feature distribution and classification neural network make the computation of \( \rho \) difficult, and the infimum is typically based on an infinite set \( \mathcal{H} \) of mapping functions. We follow the approximation method for LDM proposed in the work \cite{LDM} to estimate the LDM value for each target domain sample, which has been proven to be an asymptotically consistent and effective estimation of LDM. Specifically, we use the statistical probability of prediction disagreement by the model on target domain samples to replace the probability \( \mathbb{P} \) and construct the hypothesis set \( \mathcal{H} \) with a set finite number of variances \( \{\sigma_k^2\}_{k=1}^K \). The estimator denoted by \( L_{e}(g, \mathcal{X}_0) \) is defined as follows:
\begin{equation}
L_{e}(g, \mathcal{X}_0) := \inf_{h \in \mathcal{H}_{g,\mathcal{X}_0}^{K}} \left\{ \rho_{tu}(h, g) \overset{\Delta}{=} \frac{1}{N_{tu}} \sum_{i=1}^{N_{tu}} \mathbb{I}[h(\mathcal{X}_i) \neq g(\mathcal{X}_i)] \right\}
\end{equation}
where \( \mathbb{I} \) is an indicator function, \( \mathcal{X}_i = (e_t^i, e_v^i, x_t^i, x_v^i) \) denotes the features of the \(i\)-th sample in the feature space \(\mathcal{X}\), \( N_{tu} \) is the number of target domain unlabeled samples for approximating \( \rho \), and \( K \) is the number of sampled hypotheses for approximating \( L \). We define a set of linearly increasing variances \( \{ \sigma_k^2 \}_{k=1}^K \) (\( 0 < \sigma_k \le 1 \)) to apply Gaussian perturbations to the model weights, thereby constructing the hypothesis set.

We apply perturbations to the expert fusion network to influence its predictions by introducing Gaussian perturbations \( \mathcal{N}(\cdot, \sigma^2) \) to the weights of the last layer of each modal classifier, as the weights of the last layer critically affect the prediction results. Before perturbation, we denote the weights of the last layer of the three classifiers as \( \mathbb{W}_{cls\_t}, \mathbb{W}_{cls\_v}, \mathbb{W}_{cls\_c} \) respectively, and the expert fusion network as \( F \). After perturbation, the weights of the last layer of the three classifiers are denoted as \( \widetilde{\mathbb{W}}_{cls\_t}, \widetilde{\mathbb{W}}_{cla\_v}, \widetilde{\mathbb{W}}_{cls\_c} \), and the expert fusion network as \(\widetilde{F}\). The perturbation process and the definitions of \( F \) and \(\widetilde{F}\) are as follows:
\begin{equation}
\widetilde{\mathbb{W}}_{cls} \sim \mathcal{N}(\mathbb{W}_{cls}, \sigma^2), cls \in \{cls\_t, cls\_v, cls\_c\}
\end{equation}
\begin{equation}
    F = F(;\mathbb{W}_{cls\_t}, \mathbb{W}_{cls\_v}, \mathbb{W}_{cls\_c})
\end{equation}
\begin{equation}
    \widetilde{F} = \widetilde{F}(;\widetilde{\mathbb{W}}_{cls\_t}, \widetilde{\mathbb{W}}_{cls\_v}, \widetilde{\mathbb{W}}_{cls\_c})
\end{equation}

The perturbation with varying variances is conducted over \(K\) rounds, with \(J\) weight samplings performed during each round of perturbation, each causing slight changes to the weights of the fusion network, i.e., the decision boundary. Based on the fusion network with altered weights, we again predict all samples in the target domain and calculate the overall proportion of samples with inconsistent predictions before and after perturbation, which serves as the candidate LDM value for each update. The LDM value (i.e., \( L_{e} \)) of a sample is updated to a smaller candidate LDM only when that sample is predicted inconsistently in a certain round. Samples with smaller \( L_{e} \) values indicate that, compared to other samples, they are more susceptible to changes in the decision boundary. The formula for updating the \( L_{e} \) of samples during the entire perturbation process is expressed as follows:
\begin{equation}
    L_{e}(\mathcal{X}_i) = 
\begin{cases}
    \min(L_{e}(\mathcal{X}_i), \rho_{tu}(\widetilde{F}, F)), & \text{if}\ F(\mathcal{X}_i) \neq \widetilde{F}_{k,j}(\mathcal{X}_i) \\
    L_{e}(\mathcal{X}_i), & \text{else}
\end{cases}
\end{equation}

Here, \( \widetilde{F}_{k,j} \) denotes the fusion network after the \(j\)-th weight sampling in the \(k\)-th round of perturbation using variance \( \sigma_k \) and \( L_{e}(\mathcal{X}_i) \) is initialized to 1. After the entire perturbation process is completed, each target domain sample obtains an \( L_{e} \) value. We sort the \( L_{e} \) values of all samples in ascending order and select the top \(m k\) samples as candidate samples for the next stage of selection, where \(m\) is a hyperparameter and \( k \) is the budget for active selection in each round.

\subsection{Multi-view Diversity Calculator}

Selecting samples solely based on the uncertainty selector may result in a large overlap of similar feature distributions among the chosen samples. Therefore, we further refine the selection process by considering sample diversity to choose more informative target domain samples (Figure~\ref{fig:ADOSE}(c)). Since each modality of the news samples contains rich information, we compute sample diversity using multi-view features. Specifically, instead of directly using the input features of the classifiers, we leverage the shallow networks of each classifier (the first layer of MLP) in the fusion network as modality extraction layers. This is because we believe that the features obtained by the classifiers before the final classification stage possess stronger instance-level discriminability. We denote the shallow networks of each classifier as \( \mathrm{M}_t, \mathrm{M}_v, \mathrm{M}_c \), and the features extracted from these networks for the text, visual, and crossed modalities are \( f_t, f_v , f_c \), respectively. The formula is expressed as follows:
\begin{equation}
    f_t = \mathrm{M}_t(e_t),\  f_v = \mathrm{M}_v(e_v), \ f_c = \mathrm{M}_c([x_t, x_v])
\end{equation}

Then, we use cosine similarity to compute the inter-sample similarity matrices for each modality and take the arithmetic mean of the three similarity matrices to obtain a global similarity matrix. The global similarity matrix captures the similarity across multiple modalities of news samples, allowing for a comprehensive assessment of instance-level similarities between samples. Furthermore, we aggregate the similarity of a sample with all other samples to calculate the diversity score \( d_i \) of \(i\)-th sample, which is defined as follows (\(f_{\cdot}^i, f_{\cdot}^j\) represent the features of the \(i\)-th and \(j\)-th samples, respectively):
\begin{equation}
    d_i = \frac{1}{3N_{tu}} \sum_{j=1}^{N_{tu}} \left[cos(f_t^i, f_t^j) + cos(f_v^i, f_v^j) + cos(f_c^i, f_c^j) \right]
\end{equation}

A larger \(d_i\) indicates that the sample is more dissimilar to other samples in the feature space. Finally, we select the top \(k\) samples with the largest \(d_i\) values from the \(mk\) samples for annotation. The final selected samples integrates uncertainty estimation and diversity calculation, and is considered the most informative for facilitating domain adaptation.

\subsection{ADOSE Training}
We define the objective function for the classification prediction results of the MEFN module as \( \mathcal{L}_{ce} \), which consists of the cross-entropy loss \( \mathcal{L}_{efn} \) from the expertise fusion network and the cross-entropy loss \( \mathcal{L}_{cls} \) from multiple modality classifiers, i.e., \( \mathcal{L}_{ce} = \mathcal{L}_{efn} + \lambda_c \mathcal{L}_{cls} \) (\(\lambda_c\) is a hypermeter). Let \(N_l\) be the total number of labeled samples from all domains, with \( N_l = N_{s,1} + N_{s,2} + ... + N_{s,M} + N_{tl} \) and \( P_{\cdot}^i(y) \) be the probability of the corresponding network predicting the label \(y\) for the \(i\)-th sample. We denote the set of three classifiers as \(CLS = \{cls\_t, cls\_v, cls\_c\} \), and the detailed definitions of the above losses are given as follows:
\begin{equation}
    \mathcal{L}_{efn} = -\frac{1}{N_l} \sum_{i=1}^{N_l} \sum_{y \in \mathcal{Y}} y\log P_{mefn}^i (y) 
\end{equation}
\begin{equation}
    \mathcal{L}_{cls} = -\frac{1}{N_l} \sum_{i=1}^{N_l}\sum_{cls \in CLS} \sum_{y \in \mathcal{Y}} y\log P_{cls}^i(y)
\end{equation}

The expertise generated by multiple classifiers should exhibit as much consistency as possible on true news. In other words, we expect that the probability distributions of the classifiers on positive samples are consistent. Accordingly, the negotiation loss \( \mathcal{L}_{nego} \) between classifiers is defined based on JS divergence as follows:
\begin{equation}
\mathcal{L}_{nego} = \frac{1}{2N_{l+}} \sum_{i=1}^{N_{l+}}[\mathrm{JS}(P_{cls\_t} || P_{cls\_c}) + \mathrm{JS}(P_{cls\_v} || P_{cls\_c}) ]
\end{equation}
where \(N_{l+}\) is the total number of positive samples (i.e., real news) in the labeled dataset \( (\bigcup_{i=1}^{M} \mathcal{S}_i) \cup \mathcal{T}_l \). We introduce hyperparameters \(\lambda_a, \lambda_t\) and \(\lambda_n\) to balance the contributions of the other losses besides \( \mathcal{L}_{ce} \).The overall loss function for the ADOSE model can be formulated as:
\begin{equation}
    \mathcal{L} = \mathcal{L}_{ce} + \lambda_a \mathcal{L}_{adv} + \lambda_t \mathcal{L}_{ctr} + \lambda_n \mathcal{L}_{nego}
\end{equation}

\begin{table*}[h]
    \centering
    \renewcommand{\arraystretch}{1.0}
    \caption{Performance comparison of domain adaptation accuracy (\%) and average F1 score (\%) of different methods (UDA and ADA) with ADOSE on Pheme and Weibo datasets. \(C, S, O, F\) represent the four event domains on Pheme dataset, and \(S, E, D, H\) represent the four topic domains on Weibo dataset.}
    \resizebox{\textwidth}{!}{%
    \begin{tabular}{lccccccccc}
        \toprule
        Dataset & Setting & Method & \textit{SOF} $\to$ \textit{C} & \textit{COF} $\to$ \textit{S} & \textit{CSF} $\to$ \textit{O} & \textit{CSO} $\to$ \textit{F} & Avg. Acc & Avg. F1(fake) & Avg. F1(real) \\
        \midrule
        \multirow{8}{*}{Pheme} & \multirow{2}{*}{UDA}
        & RDCM \cite{RDCM}          & 76.58 & 63.77 & 67.53 & 88.67 & 74.13 & 43.82 & 80.38 \\
        & & ADOSE-UDA               & 75.60 & 67.71 & 74.02 & 86.79 & 76.03 & 54.54 & 80.92 \\
        \cmidrule(lr){2-10}
        & \multirow{5}{*}{ADA}
        & Detective \cite{detective}   & 77.07 & 75.59 & 74.02 & 87.73 & 78.60 & 47.30 & 83.15 \\
        & & CLUE \cite{clue}           & 76.09 & 55.90 & 57.14 & 87.73 & 69.21 & 20.16 & 79.12 \\
        & & EADA \cite{eada}           & 75.12 & 70.86 & 81.81 & 87.73 & 78.88 & 51.95 & 81.51 \\ 
        & & Entropy \cite{entropy}     & 77.56 & 73.22 & 80.51 & 88.67 & 79.99 & 63.05 & 82.87 \\
        & & \textbf{ADOSE}(ours) & \textbf{80.00} & \textbf{77.95} & \textbf{84.41} & \textbf{90.56} & \textbf{83.23}  & \textbf{64.17} & \textbf{86.29} \\         \hline
        \midrule
        Dataset & Setting & Method & \textit{EDH} $\to$ \textit{S} & \textit{SDH} $\to$ \textit{E} & \textit{SEH} $\to$ \textit{D} & \textit{SED} $\to$ \textit{H} & Avg. Acc & Avg. F1(fake) & Avg. F1(real) \\
        \midrule
        \multirow{8}{*}{Weibo} & \multirow{2}{*}{UDA}
        & RDCM \cite{RDCM}         & 75.12 & 77.31 & 77.98 & 88.09 & 79.62 & 72.27 & 79.77 \\
        & & ADOSE-UDA   & 75.72 & 78.41 & 83.48 & 84.52 & 80.53 & 75.30 & 80.95 \\
        \cmidrule(lr){2-10}
        & \multirow{5}{*}{ADA}
        & Detective \cite{detective}     & 81.14 & 82.15 & 80.73 & 89.28 & 83.32 & 77.61 & 84.47 \\
        & & CLUE \cite{clue}        & 78.13 & 81.05 & 78.89 & 87.50 & 81.39 & 74.36 & 81.51 \\
        & & EADA \cite{eada}       & 81.84 & 81.71 & 78.89 & 89.88 & 83.08 & 76.32 & 84.07 \\
        & & Entropy \cite{entropy}     & 81.44 & 81.93 & 80.73 & 91.66 & 83.94 & 78.89 & 85.15 \\
        & & \textbf{ADOSE}(ours) & \textbf{84.15} & \textbf{83.70} & \textbf{87.15} & \textbf{91.66} & \textbf{86.66} & \textbf{82.67} & \textbf{87.23} \\
        \bottomrule
    \end{tabular}
    \label{tab:perf}}
\end{table*}

\section{EXPERIMENTS}

\subsection{Experiment Setup}
\subsubsection{Datasets.}
Our model is evaluated on two real-world datasets: Pheme \cite{Pheme} and Weibo \cite{weibo}. Pheme dataset is collected from five breaking news events related to terrorist attacks. We follow the same data preprocessing methods as the benchmark \cite{RDCM} and retain four event domains with enough available text and images: Charlie Hebdo (\(C\)), Sydney Siege (\(S\)), Ferguson Unrest (\(F\)) and Ottawa Shooting (\(O\)). Weibo dataset is divided into nine topic domains by the work \cite{MMDFND}. Similar to the setup of Pheme, we remove some domains with a small number of samples and finally get four topic domains: society (\(S\)), entertainment (\(E\)), education (\(D\)) and health (\(H\)). We split the above datasets into 70\% training sets and 30\% testing sets according to each domain. To meet the cross-domain setting, during the experiments, we set three news domains from either the Pheme dataset or the Weibo dataset as source domains, with the remaining one as the target domain.

\subsubsection{Implementation Details.}
We employ TextCNN and ResNet50 as the backbone framework to extract initial text and image embeddings and map the embeddings into \(d\) dimensions, using corresponding two-layer MLPs, where \(d\) is set to 256. We adopt Adam \cite{Adam} as the optimizer with a learning rate 0.001 and weight decay of 0.0005. The sample size of each domain is set to 16 for each minibatch. Following \cite{Ada-id}, the total labeling budget \(\mathcal{B}\) is set as 10\% of target samples, which is divided into 5 selection rounds, i.e., the labeling budget in each round is \(k = \mathcal{B}/5\). The hyperparameters of the objective function are determined through experimental search as \( \lambda_c = \lambda_t = 0.5 \), \( \lambda_a = \lambda_n =0.2 \). We find that \( m=2 \) achieves the best performance on the Pheme dataset, while \( m=5 \) is optimal for the Weibo dataset.

\subsubsection{Evaluation Metrics.}
We utilize accuracy as the main evaluation metric. The average accuracy (Avg. Acc) is the mean of four accuracy values obtained from different experimental settings of target domains. We additionally employ the average F1 score for negative samples i.e., Avg. F1(fake) and for positive samples i.e., Avg. F1(real). Avg. F1(fake) represents the mean F1-value for fake news samples across four domain adaptation scenarios on the same dataset, and Avg. F1(real) is obtained similarly.

\subsubsection{Baselines.}
To conduct a comprehensive evaluation of our proposed model, we compare it with previous state-of-the-art UDA fake news detection methods and ADA approaches. UDA methods include RDCM \cite{RDCM} and ADOSE-UDA which is derived from our proposed model ADOSE by removing the process of actively selecting target samples (i.e., the LUS and MDC). For ADA methods, we choose Detective \cite{detective}, CLUE \cite{clue}, EADA \cite{eada} as well as the most commonly used active selection method, entropy sampling (Entropy) \cite{entropy}. Since the active learning method Entropy cannot be used alone in our task, we employ our MEFN module as the classification backbone for Entropy.
%
\subsection{Overall Performance}

To assess our proposed method from both domain-specific and average perspectives, we conduct comparisons between ADOSE and various domain adaptation methods (include UDA and ADA) on multimodal fake news detection. Table~\ref{tab:perf} summarizes the quantitative experiment results of our framework and baselines on the Pheme and Weibo datasets, respectively. We make the following observations: 

\begin{itemize}[topsep=5pt, leftmargin=2em]
\item In general, ADOSE achieves the best performance on almost all the adaptation scenarios compared to various baselines. In particular, our method substantially outperforms others in terms of average accuracy (Avg. Acc) of classification, achieving an improvement of \(3.24\% \sim 14.02\%\) on the Pheme dataset and \(2.72\% \sim 7.04\%\) on the Weibo dataset. 
\item Another easily observable trend is that, except for the performance of CLUE on the Pheme dataset, almost all ADA methods outperform UDA methods, where UDA methods mainly utilize the samples features to train model without accessing the sample labels in the target domain. The trend is more pronounced in the \(COF \to S\) and \(CSF \to O\) settings of Pheme and the \(EDH \to S\) and \(SDH \to E\) settings of Weibo. It suggests the effectiveness of strategic active annotation in addressing domain adaptation challenges. 
\item Evidence, energy, and entropy based on uncertainty ADA methods (\textit{e.g.}, Detective, EADA, Entropy), as well as clustering-based ADA methods (\textit{e.g.}, CLUE), generally fail to exert their inherent advantages in multimodal binary fake news detection tasks, as evidenced by lower Avg. F1(fake) values. In contrast, our Least-disagree Uncertainty selector (LUS) and Multi-view Diversity Calculator (MDC) are better adapted to the target task, with optimal performance in Avg. F1(fake) and Avg. F1(real) scores, further demonstrating the superiority of our proposed framework. More comparative results on the F1(fake) and F1(real) metrics across different scenarios are shown in Supplementary Materials C.
\end{itemize}

\subsection{Ablation Study}

To evaluate the effectiveness of each component in the ADOSE framework, we perform comparative analyses by omitting each component individually. The experimental setups are as follows: \textbf{(1) w/o MEFN:} The Modal-dependency Expertise Fusion Network module is omitted, and classification is performed by solely relying cross-modal classifier. \textbf{(2) w/o LUS:} The Least-disagree Uncertainty Selector, which quantifies the closeness of samples to the decision boundary to select uncertain target samples, is removed. \textbf{(3) w/o MDC:} The Multi-view Diversity Calculator for avoiding semantic overlap in selected samples is omitted. Table~\ref{tab:ablation} presents the results of ablation studies (More detailed ablation results are shown in Supplementary Materials D).

We observe that the original ADOSE outperforms all variants on average metrics, demonstrating the efficacy of each component. We find that the module with the greatest impact on detection accuracy is the LUS module, followed by the MEFN module, with MDC having the least impact. This aligns with our module design intentions, where LUS serves as the core selection strategy, determining the overall performance of domain adaptation, and MEFN is more important than MDC due to its role in modeling multimodal dependencies.

\begin{figure}
    \centering
    \begin{subfigure}[b]{0.4\columnwidth}
        \centering
        \hspace*{-0.6cm} 
        \begin{tikzpicture}
            \begin{axis}[
                ylabel={Acc (\%)},
                ylabel style={font=\small, at={(axis description cs:0,1)}, anchor=south, rotate=-90},
                symbolic x coords={2, 3, 4, 5, 10},
                xtick={2, 3, 4, 5, 10},
                ymin=50, ymax=95,
                ytick={50, 60, 70, 80, 90},
                legend pos=south west, 
                legend style={font=\tiny, scale=0.5, fill=white, fill opacity=0.3,
                draw opacity=0.8, text opacity=1, draw=gray!50},
                grid=major, 
                width=5.2cm, height=4.5cm 
            ]

            \addplot[
                color=blue,
                mark=triangle*,
                mark options={mark size=0.8pt},
                line width=0.8pt
            ] coordinates {
                (2, 80) (3, 78.53) (4, 82.43) (5, 81.46) (10, 79.02) 
            };
            \addlegendentry{$SOF \rightarrow C$}

            \addplot[
                color=orange,
                mark=*,
                mark options={mark size=0.8pt},
                line width=0.8pt
            ] coordinates {
                (2, 77.95) (3, 65.35) (4, 74.01) (5, 73.22) (10, 67.71)
            };
            \addlegendentry{$COF \rightarrow S$}

            \addplot[
                color=red,
                mark=square,
                mark options={mark size=0.8pt},
                line width=0.8pt
            ] coordinates {
                (2, 84.41) (3, 83.11) (4, 81.81) (5, 81.81) (10, 81.81)
            };
            \addlegendentry{$CSF \rightarrow O$}

            \addplot[
                color=teal,
                mark=star,
                mark options={mark size=0.8pt},
                line width=0.8pt
            ] coordinates {
                (2, 90.56) (3, 87.73) (4, 87.73) (5, 87.73) (10, 88.67)
            };
            \addlegendentry{$CSO \rightarrow F$}

            \end{axis}
        \end{tikzpicture}
        \caption{Pheme}
        \label{fig:alpha}
    \end{subfigure}
    \hspace{2.5em}
    \begin{subfigure}[b]{0.4\columnwidth}
        \centering
        \hspace*{-0.6cm} 
        \begin{tikzpicture}
            \begin{axis}[
                ylabel={Acc (\%)},
                ylabel style={font=\small, at={(axis description cs:0,1)}, anchor=south, rotate=-90},
                symbolic x coords={2, 3, 4, 5, 10},
                xtick={2, 3, 4, 5, 10},
                ymin=50, ymax=95,
                ytick={50, 60, 70, 80, 90},
                legend pos=south west, 
                legend style={font=\tiny, scale=0.5, fill=white, fill opacity=0.3,
                draw opacity=0.8, text opacity=1, draw=gray!50},
                grid=major, 
                width=5.2cm, height=4.5cm 
            ]

            \addplot[
                color=blue,
                mark=triangle*,
                mark options={mark size=0.8pt},
                line width=0.8pt
            ] coordinates {
                (2, 83.04) (3, 80.94) (4, 81.14) (5, 84.15) (10, 77.63)
            };
            \addlegendentry{$EDH \rightarrow S$}

            \addplot[
                color=orange,
                mark=*,
                mark options={mark size=0.8pt},
                line width=0.8pt
            ] coordinates {
                (2, 83.92) (3, 82.15) (4, 84.36) (5, 83.7) (10, 77.97)
            };
            \addlegendentry{$SDH \rightarrow E$}

            \addplot[
                color=red,
                mark=square,
                mark options={mark size=0.8pt},
                line width=0.8pt
            ] coordinates {
                (2, 82.56) (3, 80.73) (4, 80.73) (5, 87.15) (10, 81.65)
            };
            \addlegendentry{$SEH \rightarrow D$}

            \addplot[
                color=teal,
                mark=star,
                mark options={mark size=0.8pt},
                line width=0.8pt
            ] coordinates {
                (2, 91.07) (3, 90.47) (4, 89.88) (5, 91.66) (10, 88.09)
            };
            \addlegendentry{$SED \rightarrow H$}

            \end{axis}
        \end{tikzpicture}
        \caption{Weibo}
        \label{fig:beta}
    \end{subfigure}
    \caption{Accuracy trends with varying $m$ (the x-axis represents the different values of $m$ as 2, 3, 4, 5, 10).}
    \label{fig:varying_m}
\end{figure}
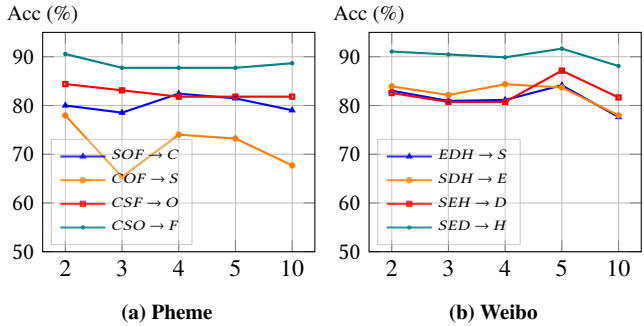

We investigate the impact of the multiplier coefficient \(m\) in the LUS module, which controls the relative influence of uncertainty and diversity in the selection strategy. A smaller \(m\) indicates that the LUS module plays a dominant role in the selection process, while a larger \(m\) reflects greater influence from the MDC module. Figure~\ref{fig:varying_m} illustrates the detection accuracy across different scenarios under varying \(m\) values on the two datasets. The figure shows that the optimal choices of \(m\) for the Pheme and Weibo datasets are 2 and 5, respectively. This phenomenon can be explained by the fact that the Pheme dataset exhibits less inter-domain differences compared to Weibo (whose domains differ significantly in topic), making uncertainty estimation more crucial than diversity measurement.

\begin{table}[H]
    \centering
    \renewcommand{\arraystretch}{1.0}
    \caption{Ablation Study Results on Pheme and Weibo Datasets.}
    \begin{tabular}{lccccc}
        \toprule
        \textbf{Dataset} & \textbf{Method} & \textbf{Avg. Acc} & \textbf{Avg. F1(fake)} & \textbf{Avg. F1(real)}  \\
        \midrule
        \multirow{4}{*}{Pheme} 
        & \textbf{ADOSE}        & \textbf{83.23}  & \textbf{64.17}  & \textbf{86.29}   \\
        & w/o MEFN              & 79.23  & 52.33  & 82.67   \\
        & w/o LUS               & 77.67  & 59.43  & 81.72   \\
        & w/o MDC               & 81.18  & 52.38  & 84.54   \\
        \midrule
        \midrule
        \multirow{4}{*}{Weibo} 
        & \textbf{ADOSE}       & \textbf{86.66}  & \textbf{82.67}  & \textbf{87.23}   \\
        & w/o MEFN             & 83.65  & 77.41  & 85.19   \\
        & w/o LUS              & 80.45  & 76.29  & 81.02   \\
        & w/o MDC              & 84.06  & 79.30  & 85.09   \\
        \bottomrule
    \end{tabular}
    \label{tab:ablation}
\end{table}

\subsection{Case Study}
\begin{figure}[htbp]
    \centering
    \begin{minipage}{0.42\columnwidth}
        \centering
        \includegraphics[width=\linewidth]{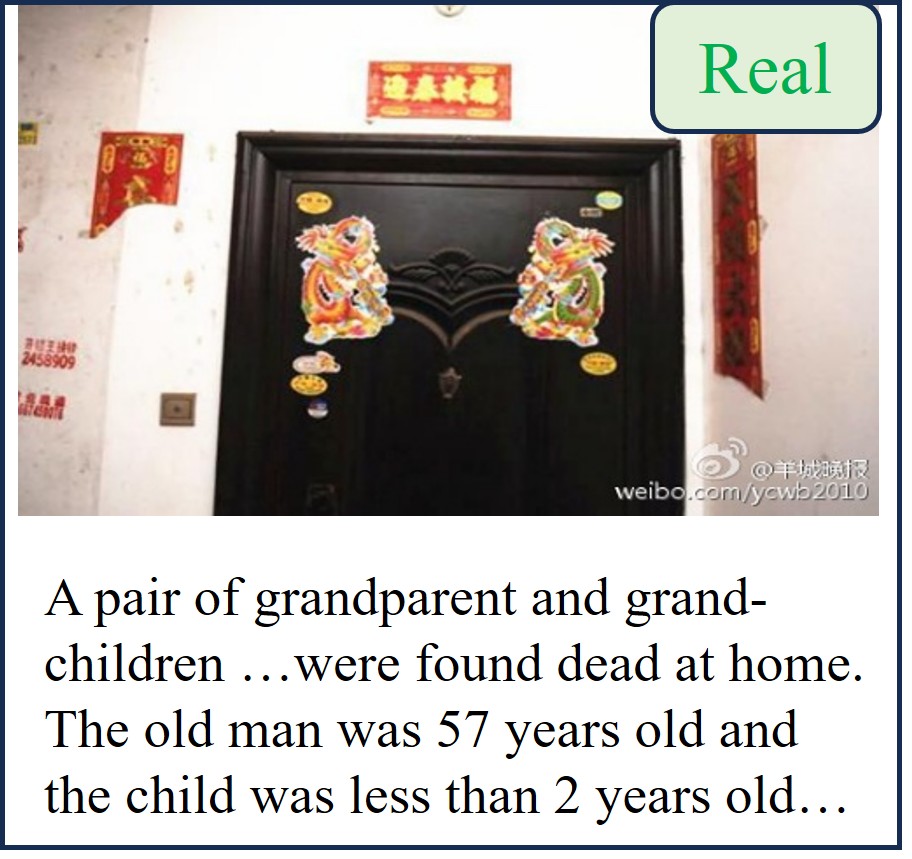}
        \subcaption{$EDH \rightarrow S$\ (Weibo)}\label{fig:sub1}
    \end{minipage}%
    \hspace{0.2cm}
    \begin{minipage}{0.42\columnwidth}
        \centering
        \includegraphics[width=\linewidth]{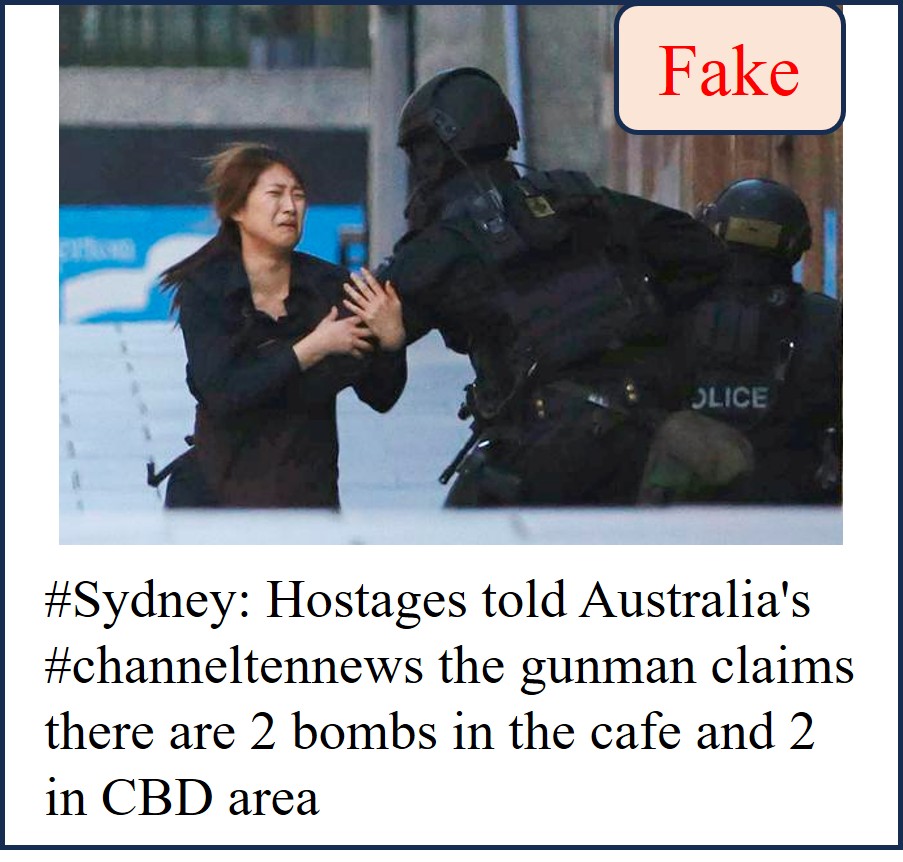}
        \subcaption{$COF \rightarrow S$ (Pheme)}\label{fig:sub2}
    \end{minipage}
    
    \vskip\baselineskip

    \begin{minipage}{0.42\columnwidth}
        \centering
        \includegraphics[width=\linewidth]{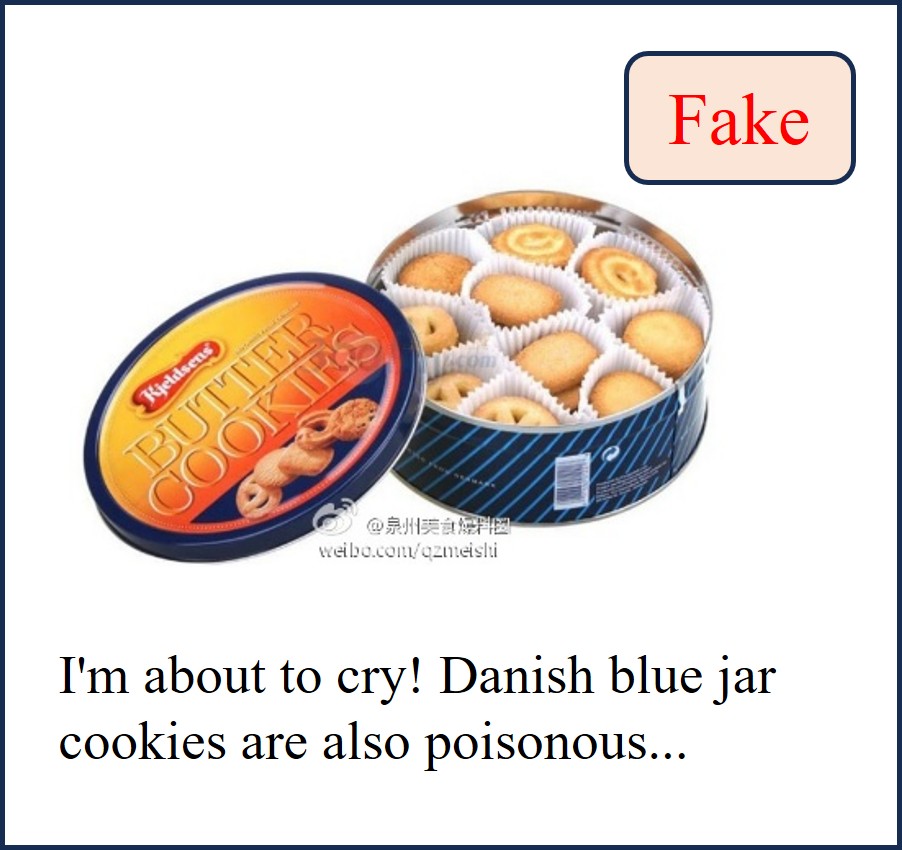}
        \subcaption{$SED \rightarrow H$ (Weibo)}\label{fig:sub3}
    \end{minipage}%
    \hspace{0.2cm}
    \begin{minipage}{0.42\columnwidth}
        \centering
        \includegraphics[width=\linewidth]{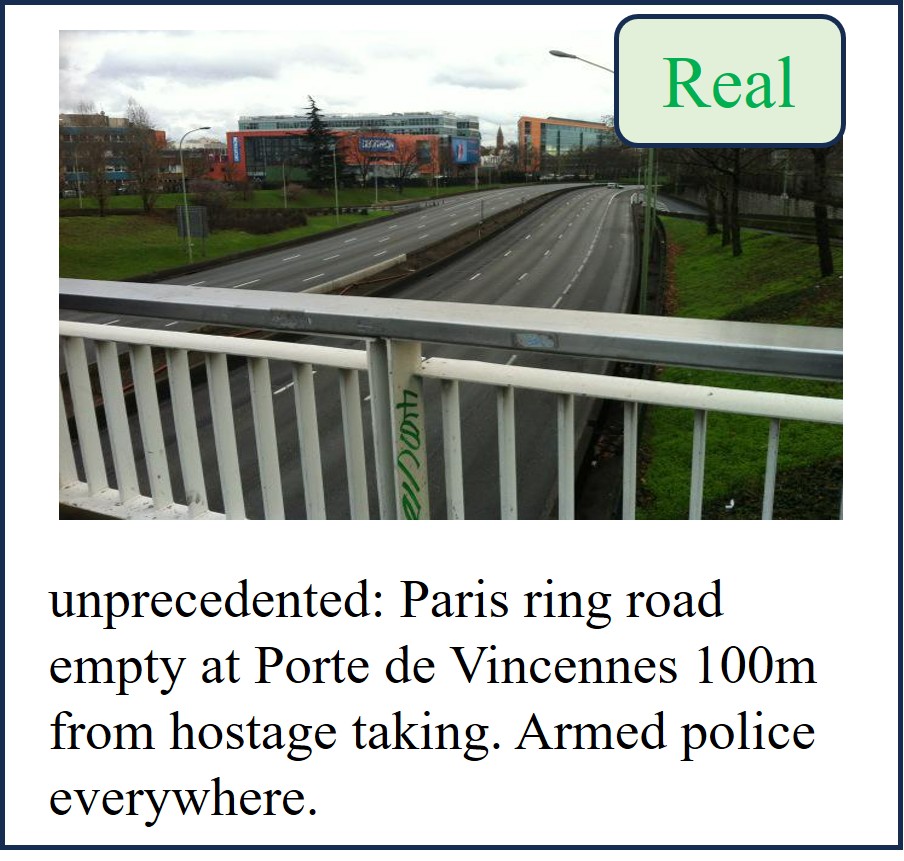}
        \subcaption{$SOF \rightarrow C$ (Pheme)}\label{fig:sub4}
    \end{minipage}
    
    \caption{The case study is conducted using target domain samples that are correctly classified by the ADOSE but incorrectly classified by most other methods.}
    \label{fig:case-study}
\end{figure}

We conduct a case study on samples that are misclassified by other methods but are detected accurately by ADOSE to justify the effectiveness of our proposed model. As shown in Figure~\ref{fig:case-study}, ADOSE can capture the textual narrative style and image color style of the target domain through training and annotating typical samples, thereby correctly classifying those confusing samples. In Figure~\ref{fig:case-study}, (a) and (b) both present objective statements of facts in the text within the target domain, with (a) appearing brighter visually and (b) appearing darker. In contrast, (c) and (d) in Figure~\ref{fig:case-study} incorporate emotional descriptions in their textual expressions, with (c) appearing to belong to warm tones and (d) to cool tones. Based on modeling domain-invariant and multimodal fusion features, our model retains these domain-specific unimodal information, enabling it to combine the advantages of active learning for better predictions in domain adaptation tasks.

\section{CONCLUSION}
In this paper, we introduce active domain adaptation into the multimodal fake news detection task to assist knowledge transfer in the target domain. Due to the varying deception patterns and domain shifts, active adaptation for multimodal fake news face two issues on the target domain: (1) Complicated modality dependencies. Distinct deception patterns often rely different modality dependencies. (2) Tailored active selection strategy. The effective selection strategy should consider both domain shift and multimodal features for news. To address these issues, we propose the ADOSE framework, which comprises the modal-dependency expertise fusion network (for the first issue) and dual active selection strategy (for the second issue). Specifically, the modal-dependency expertise fusion network separately learns intra- and inter-modal dependencies knowledge by constructing multiple experrt classifiers across domains and fuse all knowledge for fake news detection. The dual active selection strategy select most uncertain and discriminative target samples for model adaptation. ADOSE is tested on the Pheme and Weibo datasets. The experimental results conducted on the two real-world datasets prove the effectiveness of our ADOSE model. 


\renewcommand{\refname}{REFERENCES}
\bibliographystyle{ACM-Reference-Format}
\bibliography{sample-base}

\end{document}